\documentclass{article}
\usepackage{spconf,amsmath,amssymb,graphicx,booktabs,multirow,hyperref}
\hypersetup{hidelinks}
\renewcommand{\arraystretch}{0.92}

\title{When Successful Memories Mislead Embodied Agents: Memory Adaptation for Task-Conditioned Execution}
\name{\begin{tabular}{c}
Quanquan Li$^{1}$, Hongbo Zhang$^{2}$, Yihe Chi$^{1}$, Liuyang Song$^{2}$ \\
Jingyu Li$^{3}$, Yuxiang Huang$^{1}$, Hongzhen Zhang$^{1}$, Guitao Cao$^{1,*}$
\end{tabular}}
\address{$^{1}$East China Normal University, Shanghai, China \\
$^{2}$Peking University, Beijing, China \\
$^{3}$University of Science and Technology of China, Hefei, China \\
$^*$Corresponding author}

\begin{document}
\ninept
\maketitle

\begin{abstract}
Experience reuse can reduce repeated exploration in embodied agents, but a trajectory that succeeded previously may be unsuitable for the current execution context. Existing memory systems primarily optimize construction and retrieval; semantic relevance and historical success therefore remain insufficient when retrieved experience contains incompatible actions or an inappropriate level of structure. We introduce Memory Adaptation for Task-Conditioned Execution (MATE), a deterministic post-retrieval procedure that converts trajectories into execution-oriented memory. MATE removes obsolete control context, extracts condition-action-effect transitions, applies verified action normalization, selects a task-dependent representation, and serializes the result under a fixed budget without additional LLM inference. On 134 ALFWorld tasks, MATE achieves task success rates of 81.3\% and 93.3\% with Qwen2.5-14B and 72B while using approximately one-tenth of the tokens required by raw trajectories. Controlled comparisons show that verified action normalization is the principal mechanism by which MATE restores the utility of retrieved experience, supporting memory adaptation as a distinct stage between retrieval and embodied execution.
\end{abstract}


\section{Introduction}
\label{sec:intro}

Embodied language-model agents operate through a closed loop of observation, action, and environment response, and must recover from partial progress over extended interactions~\cite{choi2024d,hiagent2025}. In benchmarks such as ALFWorld and ScienceWorld, success depends not only on reasoning quality but also on whether a remembered procedure remains applicable to the current state and admissible action space~\cite{shridhar2021,rwang2022}. Reusing prior experience can reduce repeated exploration and preserve procedures that would otherwise need to be rediscovered. Shinn et al. and Zhao et al. demonstrated this benefit through remembered feedback and distilled experience~\cite{shinn2023,zhao2024}.

Subsequent work organized interaction histories as workflows, manuals, replay records, or procedural memories~\cite{wang2024,mchen2024,feng2025,fang2026}. These representations make experience easier to retrieve and reuse, but generally treat a semantically relevant trajectory as suitable evidence for the executor. This assumption is fragile in embodied interaction because the current episode may expose different objects, states, action conventions, or execution requirements. Consequently, a trajectory that was successful in its source episode can induce invalid actions and reduce performance in a task that the agent can otherwise solve.

This failure mode raises a methodological question: whether the observed degradation is caused by excessive context, irrelevant content, or incompatibility between historical actions and the current execution specification. A direct comparison between full trajectories and summaries cannot identify these effects because summarization jointly alters memory length, retained evidence, and action surface forms. We therefore require fixed retrieval and paired task outcomes, together with controls that independently manipulate memory length and action compatibility. These confounds motivate a factorized evaluation protocol in which each memory transformation is operationally explicit and causally testable.

Diagnosis alone does not determine how compatible experience should be represented. The information required by the executor depends on task structure: reactive control relies primarily on local preconditions, actions, and observable effects, whereas long-horizon planning requires constraints and dependencies spanning multiple decisions. A single compressed representation may therefore introduce task-dependent information loss. This structural heterogeneity motivates task-conditioned memory adaptation that preserves the information required by the current execution regime instead of applying a uniform summary to every retrieved trajectory.

\begin{figure*}[t!]
\centering
\includegraphics[width=0.88\textwidth,trim=4 4 4 4,clip]{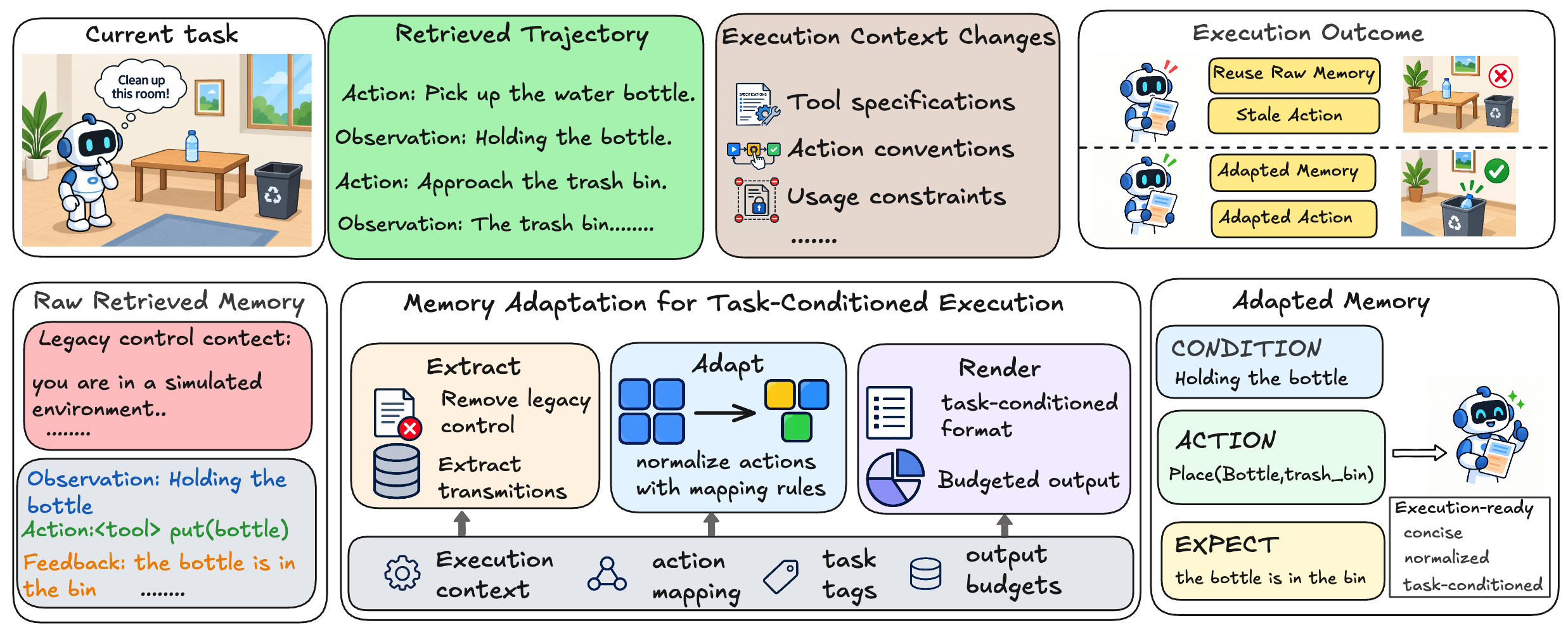}
\caption{Overview of the failure mode and MATE pipeline. Top: a retrieved trajectory may remain semantically relevant while retaining obsolete control context and actions incompatible with the current execution specification, causing direct reuse to fail. Bottom: MATE filters legacy context, extracts condition, action, and effect transitions, normalizes supported actions, and renders task-conditioned memory within a fixed budget. The resulting memory is concise, auditable, and execution-ready.}
\label{fig:mate_overview}
\end{figure*}

To address the above problems, we introduce Memory Adaptation for Task-Conditioned Execution (MATE), a deterministic post-retrieval procedure illustrated in Figure~\ref{fig:mate_overview}. Given retrieved trajectories, a current action specification, a task profile, and a budget, MATE exposes each transformation as an auditable operator: it filters obsolete control context, extracts transitions, and applies exact action-normalization rules. The task profile renders local records for reactive control or global constraints for planning, and deterministic serialization enforces the budget. Without additional LLM inference, MATE treats retrieved experience as an intermediate representation adapted to the current embodied execution regime.

Controlled ALFWorld experiments show that MATE recovers performance lost under direct trajectory reuse. It outperforms a budget-capped, action-normalized full-trajectory control but is statistically comparable to the source-matched action sequence, identifying action normalization as the principal supported mechanism. Additional structured fields show no independent success-rate gain, and fresh-memory and planning results support selective rather than universal adaptation.

Our contributions are as follows:
\begin{enumerate}
\setlength{\topsep}{1pt}
\setlength{\partopsep}{0pt}
\setlength{\itemsep}{0pt}
\setlength{\parsep}{0pt}
\item We identify and causally diagnose memory-induced negative transfer: previously successful, relevant trajectories can impair embodied execution when their action forms are unsuitable.
\item We formulate post-retrieval adaptation through MATE, a deterministic procedure for action normalization and task-conditioned rendering without additional LLM inference.
\item Fixed-retrieval controls, paired tests, ablations, and boundary evaluations identify action normalization as the main supported mechanism, quantify the efficiency of structured memory, and delimit when action sequences or full trajectories remain competitive.
\end{enumerate}

\section{Related Work}
\label{sec:related}

\subsection{Memory for Embodied and Interactive Agents}

Memory allows embodied and interactive agents to reuse information acquired across episodes while responding to the current environment~\cite{kang2025memoryos,agemem2026}. Earlier studies used verbal feedback or distilled experience to improve later attempts~\cite{shinn2023,zhao2024}. More recent methods structured trajectories as workflows, manuals, replay records, procedural memories, or hierarchical working memory~\cite{wang2024,mchen2024,feng2025,fang2026,hiagent2025,hmem2026}. Although these representations improve construction and retrieval, they generally pass selected experience to the executor without an explicit test of whether its action forms and state assumptions remain valid for the current task.

\subsection{Experience Reuse and Negative Transfer}

Experience reuse can consequently produce negative transfer. Xiong et al. documented error propagation under experience following~\cite{xiong2026}, while subsequent analyses attributed failures to conflicting advice, spurious trajectory correlations, and mismatches between transfer conditions and retrieval organization~\cite{ychen2026,tang2026,hu2026}. Environment-probing curation further showed that candidate memories may require verification before execution~\cite{suresh2026}. These findings establish that semantic relevance and historical success do not guarantee current utility. However, prior work has mainly selected, calibrated, or validated memory; it has not defined a deterministic transformation that preserves valid task structure while repairing unusable action expressions.

\subsection{Task-Conditioned Trajectory Representations}

The appropriate representation also depends on task structure. ReAct formalized the alternation between reasoning and acting, and later embodied systems separated reasoning from plan rendering~\cite{yao2023,choi2024d}. Long-horizon benchmarks such as TravelPlanner additionally require constraints that span multiple decisions~\cite{xie2024}. Local condition-action-effect records are economical for reactive control but may omit cross-step dependencies; full trajectories retain global context but also preserve irrelevant or invalid actions. Together with the compatibility failures discussed above, this representational trade-off motivates a post-retrieval adaptation stage that normalizes historical actions and renders memory according to the current task structure, rather than applying a uniform summary to every trajectory.

\section{Methods}
\label{sec:mate}

MATE operates between memory retrieval and embodied execution. It does not generate a new plan or modify the executor; instead, it deterministically transforms retrieved trajectories into an execution-oriented representation under the current task specification. We first define this adaptation objective and then detail the transformation, rendering, and serialization protocols.

\subsection{Embodied Task and Adaptation Objective}
\label{sec:task}

At interaction step $t$, an embodied agent receives observation $o_t$, selects $a_t$ from the action space defined by the current schema $\Sigma$, and observes $o_{t+1}$. For task query $q$, a fixed retriever returns ranked historical trajectories $\mathcal{R}_k(q)=\{\tau_i\}_{i=1}^{k}$, where $\tau_i=(o_1^i,a_1^i,o_2^i,\ldots,a_{T_i}^i,o_{T_i+1}^i)$. Historical success does not imply current applicability because object states, action syntax, and required planning structure may differ. We call an outcome \emph{memory-induced negative transfer} when the same executor succeeds without retrieved memory but fails after $\mathcal{R}_k(q)$ is inserted. The adaptation objective is therefore to preserve task-relevant transition or constraint structure while enforcing the current action specification and a token budget. This formulation separates memory adaptation from retrieval quality and from task-fact correctness.

\subsection{Deterministic Memory Adaptation Protocol}

MATE receives $\mathcal{R}_k(q)$, schema $\Sigma$, an exact rewrite map $\mu_{\Sigma}$, a supplied task profile $z$, and budget $B$. It produces
\begin{equation}
\label{eq:mate_pipeline}
\widetilde{\mathcal{M}}=
(\mathcal{S}_{B}\circ\mathcal{P}_{z}\circ\mathcal{N}_{\Sigma}
\circ\mathcal{E}\circ\mathcal{F})(\mathcal{R}_{k}(q)),
\end{equation}
where $\mathcal{F}$ filters obsolete control context, $\mathcal{E}$ extracts execution structure, $\mathcal{N}_{\Sigma}$ normalizes supported actions, $\mathcal{P}_{z}$ renders the task-conditioned representation, and $\mathcal{S}_{B}$ serializes it such that $|\widetilde{\mathcal{M}}|_{\mathrm{tok}}\leq B$. Equation~\ref{eq:mate_pipeline} enforces three invariants: source memories retain their retrieval order and identifiers, only exact rule matches are rewritten, and identical inputs yield identical serialized memory. Consequently, context shortening alone is not equivalent to MATE because it enforces neither action validity nor task-dependent structure.

\subsection{Transition Extraction and Action Normalization}
\label{sec:transformation}

\textbf{Context filtering.} For message-form trajectories, $\mathcal{F}$ removes the historical system or user instruction and an immediately following acknowledgement, while retaining the interaction trace. This prevents an obsolete control prompt from competing with the current instruction without deleting environment evidence.

\textbf{Transition extraction.} The extractor recognizes explicit action fields and pairs each action with the nearest preceding and following environment messages:
\begin{equation}
\label{eq:transition}
\mathcal{E}(\tau_i)=
\bigl[\langle p_j,a_j,e_j\rangle\bigr]_{j=1}^{n_i},\quad
p_j=\operatorname{prev}(a_j),\ e_j=\operatorname{next}(a_j).
\end{equation}
For serialized traces, the preceding observed effect becomes the next precondition. Whitespace is canonicalized and each precondition and effect is capped at 320 characters. Free-form thoughts may support parsing but are not copied into the resulting procedural record.

\textbf{Verified action normalization.} The normalization operator applies the supplied map only when an action exactly matches a supported source form:
\begin{equation}
\label{eq:normalization}
\widehat a_j=
\begin{cases}
\mu_{\Sigma}(a_j), & a_j\in\operatorname{dom}(\mu_{\Sigma}),\\
a_j, & \text{otherwise}.
\end{cases}
\end{equation}
Only the first case is a verified rewrite; unmatched expressions remain unchanged and auditable. Section~\ref{sec:setup} specifies how the ALFWorld rewrite map was fixed for evaluation. MATE never infers unseen action semantics.

\subsection{Task-Conditioned Rendering and Serialization}

Let $\mathcal{X}=(\mathcal{N}_{\Sigma}\circ\mathcal{E}\circ\mathcal{F})(\mathcal{R}_k(q))$ denote the normalized structure and let $c$ denote the current task context. The supplied profile $z$ selects the renderer explicitly:
\begin{equation}
\label{eq:rendering}
\mathcal{P}_{z}(\mathcal{X};c)=
\begin{cases}
\mathcal{P}_{\mathrm{react}}(\mathcal{X}), & z=\mathrm{reactive},\\
\mathcal{P}_{\mathrm{global}}(\mathcal{X};c), & z=\mathrm{global}.
\end{cases}
\end{equation}
In the first branch of Equation~\ref{eq:rendering}, \textbf{reactive rendering} converts Equation~\ref{eq:transition} into ordered precondition-action-effect records $\langle p_j,\widehat a_j,e_j\rangle$. Each record exposes the historical precondition under which an action was effective and its observed effect; the executor decides whether that precondition matches the current state. In the second branch, \textbf{global-constraint rendering} re-instantiates the current constraint state from $c$, including route, dates, party size, budget, and explicit requirements. Historical task values are discarded. A deterministic rule derives reusable constraint categories and closure order from the constraint types associated with each retrieved successful memory. For TravelPlanner, the resulting record requires transportation, accommodation, meals, attractions, and total cost to be closed before finalization. The profile is fixed at the benchmark level, with reactive rendering for embodied control and global-constraint rendering for TravelPlanner; no learned router is used.

The serializer prepends a current-interface instruction and traverses memories in retrieval order. Complete memory items are appended while the candidate remains within $B$; if no complete item fits, the serializer retains the largest line-aligned prefix of the first item within the remaining budget. The output metadata stores source, selected, dropped, and partially retained memory identifiers, token counts, actions before and after normalization, and a content hash. These records make each compiled memory reproducible and traceable to its sources. MATE has no learnable parameters and invokes no additional LLM inference. Its guarantees are limited to supplied rules and observed structure: it cannot validate an unmatched action or recover a global constraint absent from both the task specification and retrieved experience. Section~\ref{sec:boundary} evaluates these limits.

\section{Experiments}
\label{sec:setup}

\subsection{Datasets and Benchmarks}

We evaluated 134 ALFWorld tasks~\cite{shridhar2021} under the released MemP protocol with a separate pool of 252 successful training trajectories. Frozen manifests fixed retrieved identifiers and ranks. ScienceWorld~\cite{rwang2022} used 149 tasks and memory collected by the current executor. TravelPlanner~\cite{xie2024} provided 1,000 strict-Final-Pass instances for a non-embodied boundary test.

\subsection{Implementation Details}

We used a ReAct loop~\cite{yao2023} with Qwen2.5-14B-Instruct, Qwen2.5-72B-Instruct~\cite{yang2024}, and gpt-5.4-mini. Qwen runs fixed BGE-M3 retrieval~\cite{jchen2024}, queries, $k=10$, pool, and manifests. Decoding used temperature 0, seed 0, and at most 30 steps. All conditions retained the same human policy example; \emph{no retrieved memory} was one-shot.

Historical trajectories followed the legacy MemP ALFWorld action specification, whereas execution used the current specification. A frozen deterministic map normalized supported legacy actions; unmatched expressions remained unchanged.

We reused MemP's queries, pool, retriever, and harness, inserting MATE only after retrieval. MemP variants and full trajectories supplied alternative representations. Published baselines retained their original budgets. Exact McNemar tests require task-level paired outcomes, so aggregate-only comparisons are descriptive.

\subsection{Controls and Evaluation Metrics}
\label{sec:controls}

Controls used frozen retrieval. Under a 2,048-token ceiling, the raw prefix-filled control appended unchanged memories by rank and filled the remainder with a line-aligned prefix, averaging 2,004 tokens. The capped normalized-full control applied only the frozen rewrites and packed whole memories, retaining a prefix only when the first item exceeded the ceiling; it averaged 1,501/1,500 tokens at 14B/72B. The shared ceiling therefore yielded different packing and realized lengths. The action-sequence control used MATE's selected identifiers and ranks.

For success indicators $m_i\in\{0,1\}$ over $N$ tasks, we report
\begingroup
\setlength{\abovedisplayskip}{2pt}
\setlength{\belowdisplayskip}{2pt}
\[
\mathrm{SR}(\%)=\frac{100}{N}\sum_{i=1}^{N}m_i.
\]
\endgroup
We also report TravelPlanner Final Pass Rate (FPR). Paired outcomes use two-sided exact McNemar tests. Within each model scale, Holm correction is applied separately to diagnostic (MATE versus no memory and raw full), published-method (Agent Workflow Memory, ExpeL, and Reflexion), representation (capped normalized full and source-matched actions), and 14B ablation (normalization, filtering, and budget) families. Reported $p_{\mathrm{H}}$ values are adjusted within each family; boundary tests report $p_{\mathrm{raw}}$.

\subsection{Experimental Results and Analysis}
\label{sec:results}

We organize the evidence as an overall comparison, mechanism controls and ablations, and robustness and boundary tests.

\subsubsection{Main Results}
\label{sec:main_results}

\begin{table}[ht]
\caption{Success rate (\%) on 134 ALFWorld tasks.}
\label{tab:main}
\centering
\footnotesize
\renewcommand{\arraystretch}{0.95}
\begin{tabular*}{\columnwidth}{@{\extracolsep{\fill}}lrr@{}}
\toprule
Method & Qwen 14B & Qwen 72B\\
\midrule
No retrieved memory & 72.4 & 87.3\\
Full trajectories & 20.1 & 69.4\\
MemP-script & 74.6 & 90.3\\
MemP-proceduralization & 20.9 & 62.7\\
Agent Workflow Memory & 68.7 & 90.3\\
ExpeL & 15.7 & 66.4\\
Reflexion$^{*}$ & 79.9 & 91.8\\
\textbf{MATE} & \textbf{81.3} & \textbf{93.3}\\
\bottomrule
\end{tabular*}
\par\smallskip
\raggedright
$^{*}$Reflexion allows up to three attempts; MATE uses one.
\end{table}

Table~\ref{tab:main} shows that MATE recovered 61.2 and 23.9 percentage points over raw full trajectories, with paired wins/losses of 85:3 and 34:2. Relative to no retrieved memory, it gained 9.0 points at 14B ($p_{\mathrm{H}}=0.065$) and 6.0 at 72B ($p_{\mathrm{H}}=0.021$); only the latter was significant. Among the single-pass methods in Table~\ref{tab:main}, MATE achieved the highest SR at both model scales. Within the published-method family, its gains over Agent Workflow Memory and ExpeL were significant except against Agent Workflow Memory at 72B. Three-attempt Reflexion did not differ significantly from single-pass MATE.

\subsubsection{Mechanism Analysis}
\label{sec:mechanism}

\begin{table}[ht]
\caption{Cross-scale controls and 14B ablations on ALFWorld. SR: success rate (\%); lengths are mean tokens. Slash-separated lengths follow 14B/72B.}
\label{tab:controls}
\centering
\footnotesize
\renewcommand{\arraystretch}{0.92}
\setlength{\tabcolsep}{2pt}
\begin{tabular*}{\columnwidth}{@{\extracolsep{\fill}}lrrr@{}}
\toprule
\multicolumn{4}{l}{\emph{Cross-scale controls}}\\
Condition & Tokens & 14B SR & 72B SR\\
\midrule
Raw full trajectory & 15,401 & 20.1 & 69.4\\
Raw full, prefix-filled cap & 2,004 & 23.1 & 80.6\\
Action-normalized full & 15,342 & 82.1 & 88.1\\
MATE & 1,617 & 81.3 & 93.3\\
Normalized full, item-packed cap & 1,501/1,500 & 70.9 & 83.6\\
Source-matched action sequence & 290 & 79.1 & 91.8\\
\midrule
\multicolumn{4}{l}{\emph{14B ablations}}\\
\multicolumn{2}{l}{Condition} & Tokens & SR\\
\midrule
\multicolumn{2}{l}{MATE} & 1,617 & 81.3\\
\multicolumn{2}{l}{Without action normalization} & 1,619 & 51.5\\
\multicolumn{2}{l}{Without context filtering} & 1,540 & 73.1\\
\multicolumn{2}{l}{Without budget} & 7,861 & 81.3\\
\bottomrule
\end{tabular*}
\end{table}

Table~\ref{tab:controls} separates length from validity. Prefix-filling raw trajectories recovered only 3.0 and 11.2 points, whereas normalizing 19 of 877 lines raised SR to 82.1\% and 88.1\%. MATE used 10.5\% of the raw-full tokens and reached 81.3\% and 93.3\%; the uncapped normalized control remained 0.8 points higher at 14B.

\subsubsection{Ablation and Budget Control}
\label{sec:ablation}

Removing normalization reduced 14B SR from 81.3\% to 51.5\% ($p_{\mathrm{H}}=1.26\times10^{-8}$); retaining obsolete context gave 73.1\% ($p_{\mathrm{H}}=0.160$). Removing the budget preserved 81.3\% but increased mean length from 1,617 to 7,861 tokens ($p_{\mathrm{H}}=1.0$), an efficiency rather than accuracy effect. MATE exceeded the capped normalized-full control ($p_{\mathrm{H}}=0.0486/0.00195$ at 14B/72B), despite different realized lengths. Against source-matched actions, MATE-only/control-only/tied outcomes were 15/12/107 and 5/3/126 ($p_{\mathrm{H}}=0.70/0.73$). Richer transition fields therefore have no established independent accuracy gain.

\subsubsection{Robustness and Boundary Analysis}
\label{sec:boundary}

\begin{table}[ht]
\caption{Boundary results (\%). SR: task success; FPR: TravelPlanner Final Pass Rate.}
\label{tab:boundary}
\centering
\footnotesize
\renewcommand{\arraystretch}{0.95}
\setlength{\tabcolsep}{2pt}
\begin{tabular*}{\columnwidth}{@{\extracolsep{\fill}}lrrr@{}}
\toprule
Setting & No retr. & Full & MATE\\
\midrule
ALFWorld, fresh (14B) & 72.4 & 79.9 & 76.9\\
ScienceWorld (14B) & 25.5 & 38.3 & 36.9\\
\midrule
\multicolumn{4}{l}{\emph{TravelPlanner FPR by memory placement}}\\
Planner only (14B) & 0.1 & 0.8 & 0.2\\
Planner and executor (14B) & 0.1 & 1.8 & 0.1\\
Planner only (72B) & 0.3 & 5.9 & 3.6\\
Planner and executor (72B) & 0.3 & 11.3 & 6.4\\
\bottomrule
\end{tabular*}
\end{table}

Table~\ref{tab:boundary} tests source and task boundaries. With gpt-5.4-mini, raw full trajectories reduced SR from 78.4\% to 59.7\% (38 harms, 13 rescues; $p_{\mathrm{raw}}=6.2\times10^{-4}$), confirming that memory-induced negative transfer was not specific to the Qwen executors. Full trajectories and MATE were comparable with fresh ALFWorld memory and in ScienceWorld, where full memory improved over no retrieval ($p_{\mathrm{raw}}=1.9\times10^{-3}$). On TravelPlanner, global rendering exceeded local cards only directionally at 14B (0.2\% versus 0.0\% FPR), while full trajectories led MATE by up to 4.9 points, favoring selective adaptation.

\section{Discussion}
\label{sec:discussion}

The proposed method implements a two-stage memory architecture in which retrieval identifies potentially useful experience and adaptation determines the representation exposed to the executor. Rather than replacing raw trajectories, a system can retain them as provenance and expose either an adapted local representation or the fuller context required for long-horizon planning. A remaining challenge is to derive safe transformations when action specifications are incomplete or retrieved actions fall outside the verified mapping.

\section{Conclusion}
\label{sec:conclusion}

We introduced MATE, a deterministic post-retrieval method that filters obsolete context, normalizes historical actions, and renders task-conditioned memory without additional LLM inference. On ALFWorld, MATE recovered losses from raw trajectory reuse, achieved the highest single-pass SR at both model scales, and substantially reduced memory length.

\end{document}